%% file: main.tex
\documentclass[dvipsnames,format=sigconf,anonymous=false,review=false]{acmart}

\usepackage{graphicx}
\usepackage{hyperref}
\usepackage{url}
\usepackage{multirow}
\usepackage{makecell}
\usepackage{csquotes}
\input{math_commands.tex}

\input{mycommands}

\usepackage[ruled]{algorithm2e}
\usepackage{algpseudocode}
\usepackage{subcaption}
\usepackage{wrapfig}
\usepackage{lipsum}

\AtBeginDocument{%
  \providecommand\BibTeX{{%
    \normalfont B\kern-0.5em{\scshape i\kern-0.25em b}\kern-0.8em\TeX}}}

\acmDOI{10.1145/3583131.3590407} % To be updated after completing copyright process
\acmISBN{979-8-4007-0119-1/23/07} % To be updated after completing copyright process
\acmConference[GECCO '23]{The Genetic and Evolutionary Computation Conference 2023}{July 15--19, 2023}{Lisbon, Portugal}
\acmYear{2023}
\copyrightyear{2023}

\begin{document}

\title{Gradient-Informed Quality Diversity for the Illumination of Discrete Spaces}

\author{Raphaël Boige}
\authornote{Both authors contributed equally to this research.}
\affiliation{%
  \institution{InstaDeep}
  \city{Paris}
  \country{France}
}
\email{r.boige@instadeep.com}

\author{Guillaume Richard}
\authornotemark[1]
\affiliation{%
  \institution{InstaDeep}
  \city{Paris}
  \country{France}
}
\email{g.richard@instadeep.com}

\author{Jérémie Donà}
\affiliation{%
  \institution{InstaDeep}
  \city{Paris}
  \country{France}
}
\email{j.dona@instadeep.com}

\author{Antoine Cully}
\affiliation{%
  \institution{Imperial College London}
  \city{London}
  \country{United Kingdom}
}
\email{a.cully@imperial.ac.uk}

\author{Thomas Pierrot}
\affiliation{%
  \institution{InstaDeep}
  \city{Boston}
  \country{MA}
}
\email{t.pierrot@instadeep.com}

% Authors must not appear in the submitted version. They should be hidden
% as long as the \iclrfinalcopy macro remains commented out below.
% Non-anonymous submissions will be rejected without review.
\begin{abstract}

Quality Diversity (\qd) algorithms have been proposed to search for a large collection of both diverse and high-performing solutions instead of a single set of local optima. While early \qd algorithms view the objective and descriptor functions as black-box functions, novel tools have been introduced to use gradient information to accelerate the search and improve overall performance of those algorithms over continuous input spaces. However a broad range of applications involve discrete spaces, such as drug discovery or image generation. Exploring those spaces is challenging as they are combinatorially large and gradients cannot be used in the same manner as in continuous spaces. We introduce \me with a Gradient-Informed Discrete Emitter (\megide), which extends \qd optimisation with differentiable functions over discrete search spaces. \megide leverages the gradient information of the objective and descriptor functions with respect to its discrete inputs to propose gradient-informed updates that guide the search towards a diverse set of high quality solutions. We evaluate our method on challenging benchmarks including protein design and discrete latent space illumination and find that our method outperforms state-of-the-art \qd algorithms in all benchmarks.
\end{abstract}

\begin{teaserfigure}
    \centering
    \includegraphics[width=0.85\textwidth]{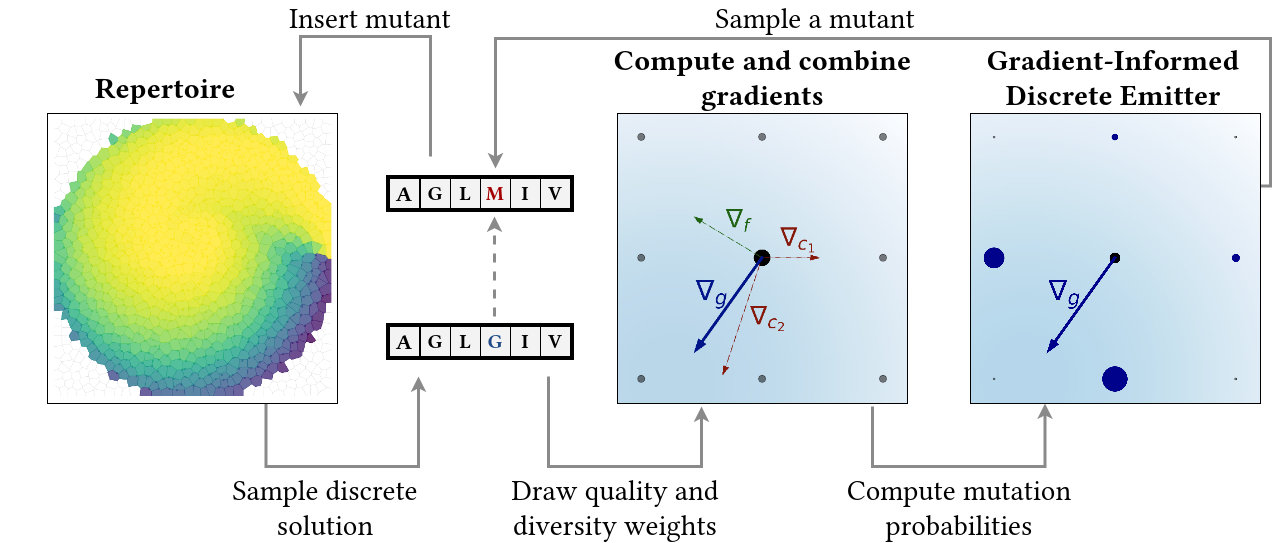}
    \caption{MAP-Elites with Gradients Informed Discrete Emitter (\megide). At each iteration, a discrete solution (here a sequence of letters from a finite vocablulary) is sampled in the repertoire. Gradients are computed over continuous fitness and descriptor functions with respect to their discrete inputs. Gradients are linearly combined to favour higher fitness and exploration of the descriptor space. Probabilities of mutation over the neighbours of the element are derived from this gradient information. Finally, a mutant is sampled according to those probabilities    and inserted back in the repertoire.}
    \label{fig:gide_main}
\end{teaserfigure}

\maketitle

\section{Introduction}

Quality-Diversity (\qd) Optimization algorithms \cite{chatzilygeroudis2021quality, cully2017quality} have changed the classical paradigm of optimization: inspired by natural evolution, the essence of \qd methods is to provide a large and diverse set of high-performing solutions rather than only the best one. This core idea showed great outcomes in different fields such as robotics \cite{cully2015robots} where it allows to learn to control robots using diverse and efficient policies or latent space exploration \cite{fontaine2021illuminating} of generative models to generate a diverse set of high quality images or video game levels. The main \qd approaches are derivative-free optimizers, such as the \me algorithm \cite{mouret2015illuminating}, a genetic algorithm that tries to find high quality solutions covering the space defined by a variety of user-defined features of interest. However, subsequent methods \cite{nilsson2021policy,fontaine2021differentiable} have shown that when the objective functions (quality) or the descriptor functions (diversity) are differentiable, using gradients of those functions can improve convergence both in terms of speed and performance compared to traditional \qd algorithms. Those methods were applied to Reinforcement Learning problems such as robotics \cite{pierrot2022diversity} or latent space illumination \cite{fontaine2021differentiable} of generative adversarial networks. 

Those applications focused on optimization over continuous variables. However many real-world applications are best framed using discrete features. For instance, generative architecture such as discrete VAEs \cite{van2017neural}; \cite{rolfe2016discrete} have shown capability to generate high-quality images \cite{razavi2019generating} and have been at the core of recent successful approaches for text-to-image generation \cite{ramesh2021zero}; \cite{gafni2022make}. A further natural application using discrete variables is protein design \cite{huang2016coming}: proteins play a role in key functions in nature and protein design allows to create new drugs or biofuels. As proteins are sequences of $20$ possible amino acids, designing them is a tremendously hard task as only a few of those possible sequences are plausible in the real world. Deep learning for biological sequences provided significant advances in several essential tasks for biological sequence, such as AlphaFold for structure prediction  \cite{Jumper2021} or language models for amino acid likelihood prediction \cite{rives2021biological} . These successes motivate the use of these tools as objective functions for protein design \cite{Anishchenko2021}.

In practice, objective functions are often written as differentiable functions of their inputs, enabling the use of gradient information of the continuous  extension if the input are discrete \cite{grathwohl2021oops}. We second this research trend and address gradient informed quality-diversity optimization over discrete inputs when objective and descriptors functions are differentiable.
We make several contributions: (i) we introduce MAP-Elites with a Gradient-Informed Discrete Emitter (\megide), a genetic algorithm where mutations are sampled thanks to a gradient informed distribution. (ii) We propose a way to select the hyperparameters of this algorithm by weighting the importance we give to this gradient information. (iii) We demonstrate the ability of our method to better illuminate discrete spaces on proteins and the discrete latent space of a generative model.

\section{Problem Definition}

In this work, we consider the problem of Quality Diversity Optimization over discrete spaces. The \qd problem assumes an objective function $f: \mathcal{X} \rightarrow \mathbb{R}$, where $\mathcal{X}$ is called search space, and $d$ descriptors $c_i: \mathcal{X} \rightarrow \mathbb{R}$, or as a single descriptor function $\mathbf{c}: \mathcal{X} \rightarrow \mathbb{R}^d$. We note $S = \mathbf{c}(\mathcal{X})$ the descriptor space formed by the range of $\mathbf{c}$. We only consider discrete spaces of dimension $m$ with $K$ categories such that $\mathcal{X} \subset \{1, ..., K\}^{m}$.
\qd algorithms of the family of the \me algorithm, discretize the descriptor space $S$ via a tessellation method. Let $\mathcal{T}$ be the tessellation of $S$ into $M$ cells $S_i$. The goal of \qd methods is to find a set of solutions $\mathbf{x}_i \in \mathcal{X}$ so that each solution $\mathbf{x}_i$ occupies a different cell $S_i$ in $\mathcal{T}$ and maximizes the objective function within that cell. The \qd objective can thus be formalized as follows:
\begin{align}
    \text{QD-Score} = \max\limits_{\mathbf{x} \in \mathcal{X}} \sum\limits_{i=1}^M f(\mathbf{x}_i), \ \text{where} \ \forall i, \ \mathbf{c}(\mathbf{x}_i) \in S_i. 
\end{align}

We consider here that both the objective and descriptor functions are actually defined on real values $\mathbb{R}$ and restricted to the discrete inputs $\mathcal{X}$. We also consider them to be first-order differentiable, hence for any input $x \in \mathcal{X}$, we can compute gradients $\nabla_x f(x)$ and $\nabla_x c_i(x)$. 

\section{MAP-Elites with Gradient Proposal Emitter}
\label{sec:me-gide}
\input{gide}

\subsection{MAP-Elites and Differentiable MAP-Elites}

\me is a \qd method that discretizes the space of possible descriptors into a repertoire (also called an archive) via a tessellation method. Its goal is to fill each cell of this repertoire with the highest performing individuals. To do so, \me first initializes a repertoire over the BD space and second initializes random solutions, evaluates them and inserts them in the repertoire. Then successive iterations are performed: (i) select and copy solutions uniformly over the repertoire (ii) mutate the copies to create new solution candidates (iii) evaluate the candidates to determine their descriptor and objective value (iv) find the corresponding cells in the tessellation (v) if the cell is empty, introduce the candidate and otherwise replace the solution already in the cell by the new candidate if it has a greater objective function value. When using real variables, the most popular choice for the mutation operator is the Iso+LineDD \cite{vassiliades2018discovering} that mixes two Gaussian perturbations. When using discrete variables, mutations are generally defined as point mutation, \textit{ie} select a random position and flip its value from the current one to a new one. 

To incorporate gradients into \me algorithms, \citep{fontaine2021differentiable} introduced MAP-Elites with a Gradient Arborescence (MEGA) a novel way to use gradient information to guide the mutations. First, the authors propose a novel objective function to encompass both quality and diversity: $g(x) = |w_0| f(x) + \sum_{i=1}^d w_i c_i(x)$, where $w_i \sim \mathcal{N}(0, \sigma_g I) \text{ } \forall i \in \{0, ..., d\}$. In one version of their algorithm, \omgmega, authors simply extend \me by mutating selected element of the repertoires $x$ via $x' = x + \nabla_x g(x)$ where $w_i$ are sampled at each iteration and for each element in the batch and $\sigma_g$ acts similarly to a learning rate. Indeed, maximizing $g$ will direct the mutations towards higher fitness and different directions of the descriptor space thanks to the randomness introduced by sampling the $w_i$.

\subsection{From Gradients to the Gradient Informed Discrete Emitter}
\input{me_gide}

As this approach has proven effective on several tasks, we follow the previous formulation by trying to maximize $g$. In this case at a given iteration, for a given sampled $x \in \mathcal{X}$ and given sampled coefficients $c_{0:...:d}$, our emitter should ideally find a neighbour $x' \in \mathcal{X}$ that maximizes the following:
\begin{equation}
    x' = \argmax_{z \in \mathcal{B}_\tau(x)} \delta(z) = g(z) - g(x)
\end{equation}
where $\mathcal{B}_\tau(x)$ is the Hamming ball of size $\tau$ around $x$.

However, for a given variable $x \in \{1, ..., K\}^m$, even for $\tau=1$ the cardinality of this Hamming ball is $mK-1$. Hence finding the optimal $x'$ requires $mK-1$ evaluations of $g$. Doing it at each step would be too expensive but since those differences are local, they can be approximated using gradient information. Following \cite{grathwohl2021oops} we use Taylor-series approximation to estimate the local differences at first-order: 
\begin{equation}
    \delta_{ik} = g(x^{(i,k)}) - g(x) \simeq \nabla_{x} g(x)_{ik} - x_i^T \nabla_x g(x)_i \textbf{ = } \tilde{\delta}_{ik}
\label{eq:appx_diff}
\end{equation}

where $x^{(i,k)}$ differs from $x$ by a flip on position $1 \leq i \leq m$ from the current value of $x_i$ to $k \in \{1, ..., K\}$. The previous formulation works in the case where $\tau=1$ but similar approximations can be derived for larger window sizes at the expense of the approximation's quality.

Using those approximations, we could straightforwardly compute a local maximum of $g$ by taking the argmax of $\tilde \delta_{ik}$ at each step. In order to encourage exploration, we use those approximate local differences to create a proposal distribution over the neighbours of $x$. We second traditional sampling technics such as Metropolis Hastings and define the flip distribution $P$ defined by the probability $p(x^{(i, k)} | x)$ of mutating from $x$ to $x^{(i, k)}$ as 
\begin{equation}
    p(x^{(i, k)} | x) \propto e^\frac{\tilde{\delta}_{ik}}{T}
    \label{eq: probsprop}
\end{equation}

where $T>0$ is a temperature parameter. Finally, in order to sample flips at position $i$ from $x_i$ to $k$, we normalize those probabilities by computing a softmax over every possible flip ($mK$ possibilities):

\begin{equation}
    p(x^{(i,k)}| x) = \frac{e^{\tilde{\delta}_{ik}/T}}{\sum\limits_{j,l=1}^{m, K} e^{\tilde{\delta}_{jl}/T}}
\label{eq:probs}
\end{equation}

As every approximated difference $\tilde{\delta}_{ik}$ can be computed in a single gradient evaluation $\nabla_x g(x)$, those probabilities can be efficiently computed. Formally, our Gradient Informed Discrete Emitter (\gide) receives a candidate $x$, computes $\tilde{\delta}_{ik}$ using Equation~\ref{eq:appx_diff}, then computes mutation probabilities using Equation~\ref{eq:probs} and finally samples a mutated $x^{(i, k)}$ using these probabilities. We summarize this procedure in Algorithm~\ref{alg:gide}. Using this emitter, we design our main algorithm: MAP-Elites with a Gradient Informed Discrete Emitter (\textbf{\megide}). The general procedure follows the one of \omgmega: we first initialize a repertoire of behaviour descriptors. Then at each iteration, the following operations are performed: (i) select a batch of solutions uniformly over the repertoire (ii) sample mutants using our \gide (iii) evaluate the candidates and add them in the repertoire if their fitness is higher than the existing solution in the cell. This procedure is described in Algorithm~\ref{alg:pseudocode} with an additional step to control the strength of the gradient guidance defined in the following paragraph.

\subsection{Controlling the Gradient Guidance with a Target Entropy}
\begin{figure}[h!]
    \includegraphics[width=0.48\textwidth]{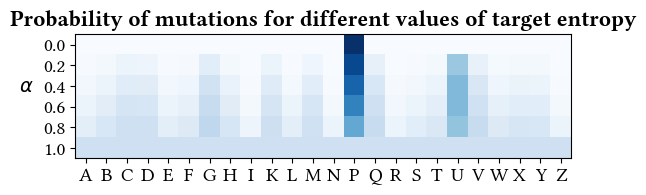}
    \caption{Probability of mutation for one position in a protein sequence, with the temperature parameter being controlled by the target entropy parametrized as $H_{target}=\alpha H_{max}$. Higher target entropy values are associated with smoother flip probabilities.}
    \label{fig:p_target_entropy}

\end{figure}

\begin{table*}[t!]
    \centering
    \begin{tabular}{lccccc}
    \toprule
     & \textbf{Ras protein family} & \textbf{Binarized digits} & \textbf{Discrete LSI - 1} & \textbf{Discrete LSI - 2} \\
    \midrule
    
    \textbf{Genotype}& \makecell{A protein sequence \\of length 184}& \makecell{A binary image \\ of dimension $28\times28$} & \makecell{A latent code \\ of length 1024}  &  \makecell{A latent code \\ of length 1024} \\
    \hline
    \textbf{Search space}& $\{0, ..., 20\}^{184}$ & $\{0, 1\}^{28\times28}$& $\{0, ..., 512\}^{1024}$ & $\{0, ..., 512\}^{1024}$  \\ 
    \hline
    \textbf{Score} & \makecell{Protein sequence likelihood \\ given by an ESM2 model}& \makecell{Image likelihood \\given by a RBM model}& \makecell{ CLIP similarity score with \\the prompt: \textquote{\textit{a labrador}}}& \makecell{ CLIP similarity score with \\the prompt: \textquote{\textit{a truck}}}\\

    \hline
    \textbf{Descriptors} &\makecell{PCA-projected embeddings\\of ESM2 hidden layer\\ ($d=5$)} & \makecell{PCA-projected embeddings\\of RBM hidden layer \\ ($d=20$)}&\makecell{CLIP scores with prompts:\\ \textquote{\textit{a dog with long hair}}, \\ \textquote{\textit{a white puppy}} ($d=2$)} & \makecell{CLIP scores with prompts: \\\textquote{\textit{a blue truck}},\\ \textit{\textquote{a red truck}} ($d=2$)} \\

    \hline
    \textbf{Centroids} &  \makecell{K-Means centroids\\ (30000 cells)}  & \makecell{K-Means centroids\\ (10000 cells)} & \makecell{Voronoi tesselation \\ (30000 cells)} &  \makecell{Voronoi tesselation \\ (30000 cells)} \\
    \bottomrule

    \vspace*{0.08cm}
    \end{tabular}
    \caption{Experimental setup for each domain: search space, scoring functions, descriptors functions and repertoire used.
    \label{tab:recap}}

\end{table*}

The temperature parameter $T$ controls the shape of the flip distribution $P$, as it tempers the softmax operator applied over the estimates $\tilde \delta_{ik}$ according to Equation~\ref{eq: probsprop}. Tuning this temperature parameter boils down to choosing whether we prefer to greedily follow the highest-gradient direction ($T \rightarrow 0$) or to sample at random a flip direction ($T \rightarrow \infty$). We propose to frame this problem as an exploration-exploitation trade-off, where we suggest to find an optimal temperature parameter $T = T^*$ such that (1) enough importance is given to the most promising directions, i.e. $T$ is not too large (2) the improvement-direction estimate is sufficiently peaky to differ from a uniform distribution, i.e. $T$ is not too small. However, this optimal temperature parameter value $T$ depends on the dimension of the search space, thus requiring a careful parameter tuning for each domain. Additionally, it also depends on the gradient landscape of the scoring and descriptors functions, which removes any guarantee that a constant value of $T$ could be optimal throughout the optimization process. 

To ease the search of the numerical value of $T$, we propose to dynamically adjust it so that the flip distribution matches, in average, a given target Shannon entropy value $H_\text{target}$. Given that we know the analytical form of the flip probability distribution $P$ thanks to Equation~\ref{eq:probs}, we can express the Shannon entropy as a function of the temperature.
\begin{equation}
    H(P) = - \sum_{i,k=1}^{m,K} p(x^{(i,k)}| x) \log(p(x^{(i,k)}| x)) = h(T)
\end{equation}
 For a given entropy target $H_\text{target}$ we can solve the equation $h(T) = H_\text{target}$ for the value of $T$, and dynamically adjust it during the optimization process. The intuition behind using a Shannon entropy target to parametrize the tempered-softmax distribution over $\tilde\delta_{ik}$ is to provide a more interpretable and consistent way to control the exploration-exploitation trade-off, by linking the temperature parameter to a measure of the randomness or uncertainty of the system. Similar Shannon entropy constraints have been previously used in the Reinforcement Learning literature \cite{haarnoja2018soft}.

In practice, at each \gide update, we use a numerical first-order solver to update the value of the temperature parameter $T$, so that we ensure the entropy target is always matched on average. In Figure~\ref{fig:p_target_entropy}, we illustrate how setting the value of $H_\text{target}$ affects the proposal distribution for the same candidate. As the entropy is bounded by $[0, H_{max}]$ with $H_{max}=\log(mk)$, this allows GIDE to be parameterized by a single hyper-parameter $\alpha \in [0, 1]$ and to set $H_\text{target} = \alpha H_{max}$.

\section{Experiments}

\subsection{Settings}
\begin{figure*}[t!]
    \centering
    \includegraphics[width=0.9\textwidth]{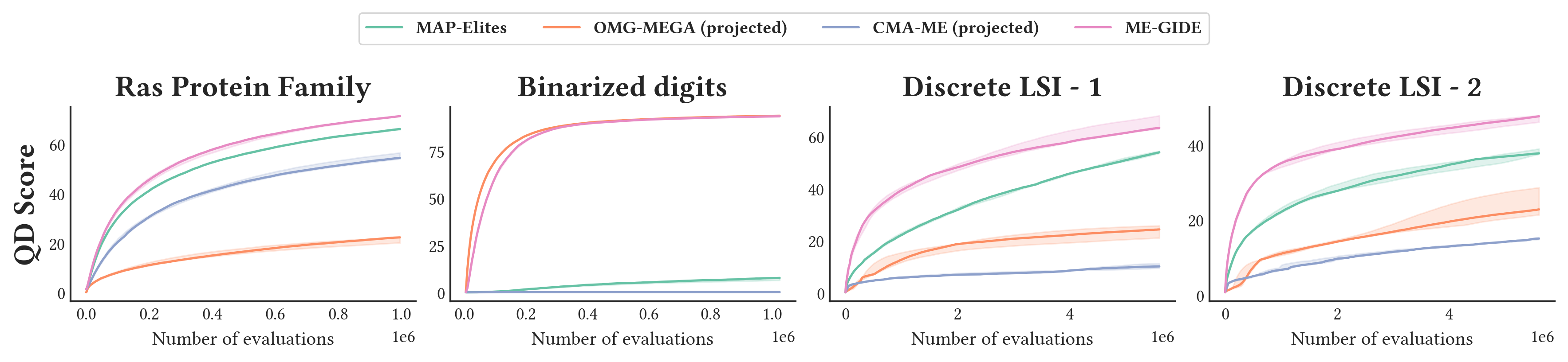}
    \caption{\qd-score evolution on the different domains (median and interquartile range over five seeds). \megide outperforms \me and \cmame on every experiment and \omgmega on every experiment except for Binarized Digits. Only the solution corresponding to the best set of hyperparameters are shown for all methods.}
    \label{fig:main_results}
    \vspace{-0.1cm}
\end{figure*}

We conduct experiments on different benchmarks to assess the performance of \megide. Namely, we experiment on three challenging applications: illuminating a family of proteins, generating diverse digits images by searching a binary pixel space and illuminating the discrete latent space of a generative model. We summarize domain characteristics in Table~\ref{tab:recap}.

\paragraph{Design in the Ras Protein Family} As proteins are involved in numerous biological phenomena, including every task of cellular life, designing them has been a longstanding issue for biologists. Relying exclusively on wet-lab experiments to iteratively select designs is both costly and time-consuming so practitioners have turned to in-silico design as a preliminary step. The simplest way to describe a protein is through its primary structure: a natural protein is a sequence of $20$ basic chemical elements called amino-acids. Designing a protein is generally framed as finding the sequence of amino acids that maximizes a score representing a property of a protein, such as its ability of binding to a virus or its stability. It is a tedious task as the search space grows exponentially with the length of the protein of interest, for instance designing a protein of $100$ residues operates over a space of $20^{100}$ elements. In our experiment, we aim to redesign members of the \textquote{Rat sarcoma virus} (\textit{Ras}) protein family (extracted from PFAM database \citep{mistry2021pfam}), involved in cellular signal transduction and playing a role in cancer development \cite{bos1989ras}. We follow a common objective in computational protein design \cite{huang2016coming}, that focuses on maximizing the stability of a family of protein. While many methods use biophysical simulations to estimate this stability \cite{desmet1992dead}, we follow recent trends \cite{anishchenko2021novo} and use a pre-trained neural network to approximate it. It is worth noting that other works use neural networks and gradient ascent to design proteins but rely on a continuous approximation of proteins \cite{castro2022transformer}.

For a given sequence, our objective function is defined as a proxy of the log-likelihood provided by a pre-trained protein language model: ESM2 \cite{Lin2022, rives2021biological}. Indeed, the log-likelihood has been shown to strongly correlate with the stability of a protein and can even be used to predict the effect of these mutations \cite{hopf2017mutation, riesselman2018deep}. We use an unsupervised procedure to create descriptors similar to AURORA \cite{grillotti2022unsupervised}: we sample 1,000,000 proteins from the Uniref50 \footnote{\url{https://www.uniprot.org/help/uniref}} database, compute their $640$-dimensional embeddings using ESM2 at the $20^{th}$ layer and perform a PCA on those embeddings to extract $5$ components. Finally we use a $K$-Means algorithm with $K=30000$ on these projected embeddings to create the tesselation with $30,000$ cells. The descriptors of a sequence is the corresponding projected embedding of this sequence and is mapped to its corresponding cell by finding the closest centroid. In practice, we use the $120,000$ elements of this family at initialization.

\paragraph{Binarized Digits}

To further illustrate the interest of our method, we design an experiment on binary data consisting in generating diverse MNIST digits with \qd methods. Given our fitness and descriptor functions, we aim to find out how \qd methods compare in generating diverse digits by directly searching over the image space. We define the fitness using an energy-based model trained on MNIST data. Following \cite{grathwohl2021oops}, we train a restricted Boltzmann machine (RBM) \cite{hinton2012practical} on the binarized $28\times28$ images, we refer the reader to the Appendix~\ref{appendix:exp_details} for further details.  To obtain descriptors we define a diversity space by embedding the MNIST dataset into the hidden layer of the RBM and then by computing a PCA over these embeddings. Finally, descriptors are defined as projections over the $\text{top-}d$ components of this PCA with $d=20$. To compute the centroids, we again use a $K$-Means algorithm with $K=10000$ on the projected embeddings. We display in Appendix~\ref{appendix:mnist_vizu} how this descriptor space spans the different digits' classes. In our experiments, we initialize our images with uniformly-distributed images from the $28\times28$ binary-images space.

\begin{figure*}[t!]
    \centering
    \includegraphics[width=0.85\textwidth]{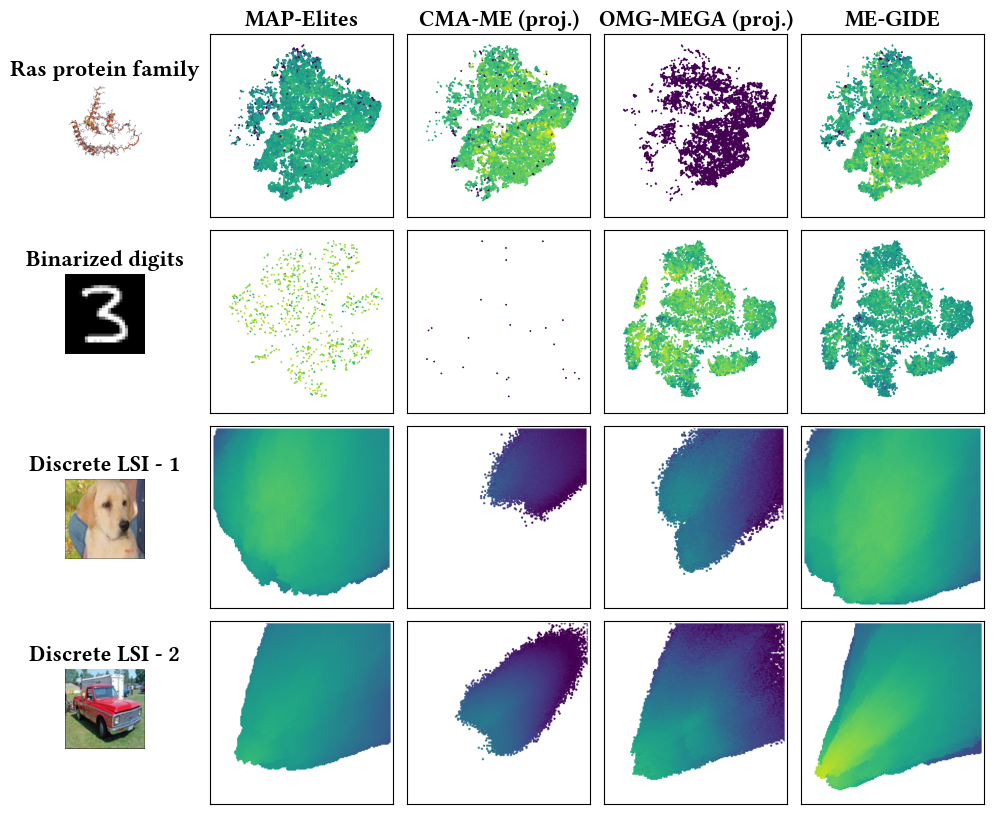}
   \caption{Repertoires for each domain (line) and method (column). The repertoires of the protein domain (resp. binarized digits domain) have been projected using t-SNE as the original one is 5-dimensional (resp. 20-dimensional). Each point in the repertoire corresponds to a solution, where dark blue corresponds to a low fitness and yellow to a high one. }
    \label{fig:repertoires}
\end{figure*}

\paragraph{Discrete Latent Space Illumination (LSI)}

Recent works \cite{fontaine2021illuminating}, \cite{fontaine2021differentiable} have proposed the latent space illumination of generative models (LSI) problem as a benchmark for \qd methods. It consists in searching the latent space of a large generative model for latent codes that can generate diverse images. We follow the setting of \cite{fontaine2021differentiable} using as generative model a discrete latent space VAE instead of a StyleGan. Since our work is focused on discrete variables, we choose to illuminate the latent space of a discrete VAE instead of a Style-GAN. Specifically, we selected a vector-quantized auto-encoder (VQ-VAE). We followed the recommendations of the seminal paper \cite{van2017neural} to design and train the VQ-VAE on ImageNet \cite{deng2009imagenet} data: we use 3 convolutional layers for the encoder and 3 layers for the decoder, a latent vector of size $32 \times 32 = 1024$
and a codebook size of $512$. Thus, each genotype for this experiment is a code of length $1024$ from the VQ-VAE latent space. that can take $512$ values, resulting in a seach space of size $512^{1024}$. We refer the reader to the Appendix~\ref{appendix:vq_vae_training} for the VQ-VAE model's details, and the training procedure.

We evaluate the solutions by decoding the proposed discrete latent codes using the pretrained VQ-VAE. We follow previous works in using the CLIP score to measure the similarity between the image and a given prompt. Specifically, we compute one CLIP score for the fitness of the solution and two CLIP scores for the descriptors. We have experimented two settings, composed of two independent sets of prompts. The first one aims to generate labrador images, the fitness is defined with the prompt \textit{\textquote{A labrador}} while the descriptor functions are defined as \textit{\textquote{A white puppy} }and\textit{ \textquote{A dog with long hair}}. The second set of prompt searches for truck images, the fitness prompt is \textit{\textquote{A truck} }and descriptor prompts are \textit{\textquote{A red truck}}, \textit{\textquote{A blue truck}}. Further experiment details were deferred to Appendix~\ref{appendix:exp_details}. After they are evaluated, solutions are inserted into a CVT \cite{vassiliades2017using} repertoire with $30,000$ cells. In all our experiments, latent codes were initialized from a random uniform distribution.

\paragraph{Experiment Setup}

In addition to our \textbf{\megide} algorithm, we compare with three main baselines. The first baseline, \textbf{\me} stands for the original \textbf{\me} algorithm with completely random one-point mutations: at each step, for each sample solution, the sequence is modified at one position to a uniformly sampled possible value. Searching via random mutations is the most popular solution when no prior knowledge is assumed over the structure of a solution. The second gradient-free baseline is MAP-Elites with Covariance Matrix Adaptation (\cmame) \cite{fontaine2021illuminating}, an extension of \me that uses CMA as a natural approximation for the gradient and has been shown competitive on many \qd tasks. Since \cmame has been designed for optimization over continuous spaces, we project back the yielded solutions to the closest points in the discrete space, we hence call this baseline \cmamep. Finally, our third baseline \omgmegap uses the same projection mechanism applied on the gradient-based method Objective and Measure Gradient MAP-Elites via Gradient Arborescence (\omgmega). The use of this baseline is natural since \omgmega can be considered as the continuous counterpart of \megide. We give pseudo-codes for all baselines in Appendix~\ref{app:baselines}. It is worth noting that we base our approach on \omgmega rather than \cmamega, the main algorithm proposed for Differentiable Quality Diversity \cite{fontaine2021differentiable} as on our benchmarks, \cmamega was under-performing compared to \omgmega. This can be explained by the fact that CMA updates are based on smoothness assumptions of the gradient that is broken by the projection step to keep solution in the discrete space.  We implement our method based on the Jax framework and use the QDax open source library \cite{lim2022accelerated, chalumeau2023qdax} for the baselines.

We run every experiment over five different seeds for each method.
 Note that in the protein experiment, all methods are combined with crossover where half of the sampled solutions are modified with the method-specific emitter and the other half with crossover. We stop the illumination after evaluating $1,000,000$ candidates on the protein and binarized digits experiment and $5,000,000$ evaluations for the discrete LSI experiments. We provide more details about the settings in Appendix~\ref{appendix:exp_details}.

\subsection{Results and Analysis}

\paragraph{Performance Analysis}Figure~\ref{fig:main_results} summarizes the results of our experiments: we display the median \qd score and the interquartile range over the five random seeds for every benchmark. In the Ras and Discrete LSI experiments, \textbf{\megide} outperforms other methods, since it converges in fewer iterations to higher performances. We use a Wilcoxon signed-rank test \cite{wilcoxon1992individual} to compare the distribution of the scores obtained over different seeds where the null hypothesis is that the obtained scores have the same median. It shows that \textbf{\megide} constantly outperforms \textbf{\me} and \cmamep in the \qd score with p-values lower than $0.05$ in every experiment and outperforms \textbf{\omgmegap} on every experiment except Binarized Digits. 

\paragraph{Entropy ablations}For clarity, we only plot the results corresponding for the best set of hyperparameters for each baseline, although we ran an hyperparameter search for each method with comparable budget as detailed in Appendix~\ref{appendix:exp_details}. Our target entropy procedure allows an easier choice of hyperparameters for \textbf{\megide} as we only need to choose $\alpha$ within a natural range $[0, 1]$. In our experiment, we try for four values for $\alpha \in \{0.2, 0.4, 0.6, 0.8\}$, we don't test $\alpha=1$ as it corresponds to standard \me with random uniform mutations. While we display the results for the best value of $\alpha$ in Figure~\ref{fig:main_results}, the conclusions are similar for other values of $\alpha$ as detailed in Appendix~\ref{appendix:full_results}.

\begin{figure}[t!]
    \includegraphics[width=0.35\textwidth]{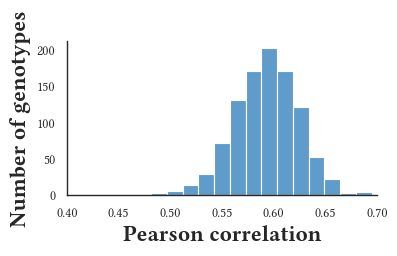}
    \caption{Pearson correlation between the true improvement $\delta(x')=g'(x)-g(x)$ and GIDE estimate $\tilde{\delta}(x')$. We sampled 1000 random latent codes from the \textit{Discrete LSI - 1} domain, we evaluated the true improvement by enumerating all the possible neighbors and we computed the correlation with the GIDE estimate. We display the distribution of theses correlations across the 1000 different latent codes.}
    \label{fig:correlation}
    \vspace{-0.4cm}
\end{figure}

\paragraph{Difficulty of gradient-informed optimization}Interestingly, \textbf{\omgmega} under-performs compared to purely random mutations for any set of hyperparameters on both protein and Discrete LSI experiments. This means performing gradient descent similarly as in the continuous case is perilous when using discrete variables as it will likely yield unfeasible solutions with no guarantee that the closest feasible point is a viable solution. As stated in Section~\ref{sec:me-gide}, our method relies on the fact that our gradient approximation $\tilde{\delta}$ is a good surrogate for the true improvement $\delta(x') = g(x') - g(x)$. This approximation relies on a smoothness assumption of those functions and the fact that neighbours stay "close" in the continuous space. We validate this assumption via the following procedure: (i) randomly sample some points $x_i$ (ii) compute for each of their neighbour $x_i'$ both $\delta(x_i')$ and $\tilde{\delta}(x_i')$ ($\tilde{\delta}$ only needs to be computed once) and (iii) compute the correlation $\rho_\delta(x)$ between those two quantities. For instance, we display the distribution of correlations for $1,000$ elements for the Discrete LSI-1 experiment in Figure~\ref{fig:correlation}, where we find an average correlation $\rho = 0.59$, which means that even though the step between $x^{(i,k)}$ and $x$ might be large, gradient information is indeed relevant. 

\begin{figure}[t!]
    %\vspace*{-0.4cm}
    \includegraphics[width=0.4\textwidth]{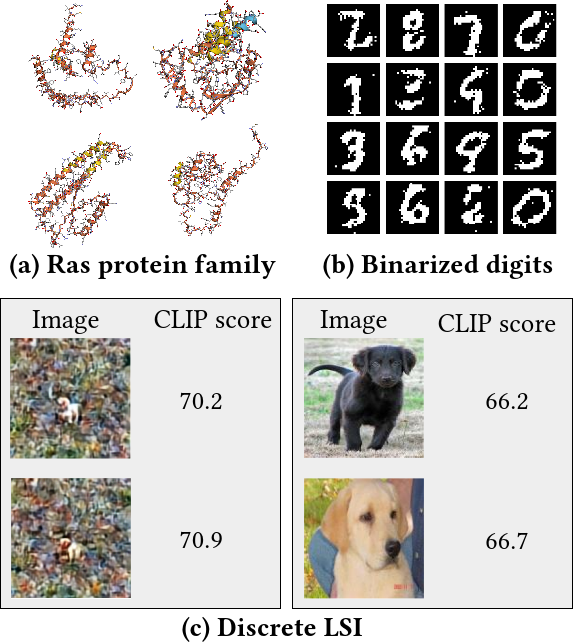}
    \caption{Illustration of the obtained solutions on the 3 domains. \textbf{(a)} \textit{Ras} proteins designed by \megide, visualized using the prediction of AlphaFold. One can see diverse patterns in the secondary structure, while still retaining high estimated stability. \textbf{(b)} Samples from \megide final repertoire on the Binarized Digits experiment. We downsample the $10,000$ elements in the repertoire by clustering its centroids to $16$ to re-create a repertoire with $16$ cells, projecting the obtained images in the downsampled repertoire. \textbf{(c)} Samples found by \megide on the "Labrador" prompt against highest scoring images from ImageNet dataset. While the images found by \megide are not natural images, they achieve a higher score to be a "labrador" according to CLIP.}
    \label{fig:domains}
    \vspace*{-0.3cm}
\end{figure}

\paragraph{Qualitative analysis}
We display some proteins of the final repertoire obtained by \megide in Figure~\ref{fig:domains}. To visualize the 3D structure of those proteins, we use AlphaFold2 \cite{jumper2021highly}. One can qualitatively see that \megide manages to extract proteins with various secondary and tertiary structures and that those proteins have high confidence score according to AlphaFold. We also conduct further quantitative validation on the protein experiment of our method \megide against \me which is the best alternative baseline in this experiment. We sub-sample our repertoire by recomputing larger cells to obtain a repertoire with $300$ cells. Then, we insert the original repertoire into this smaller one to obtain the most fit solutions in each region. We first analyze the diversity in the sequence space, defined as the average edit distance between obtained solutions. We find that \textbf{\megide} reaches an average distance of $106.9$ whereas \textbf{\me} gets $97.2$. We also use \textit{S4Pred} \cite{moffat2021increasing} to predict the secondary structure of each protein and use the average edit distance between secondary structures. \textbf{\megide} outperforms \textbf{\me}, obtaining an average distance of $44.2$ against $34.9$, and finds more diverse structures while also reaching an overall higher average fitness.

In Figure~\ref{fig:domains}, we also display some samples found by \megide on the binarized MNIST experiment. One can see that \megide is able to find diverse images that resemble MNIST images and covers every digit. We also visualize top-performing images from the repertoire learned by \textbf{\megide} as well as natural images from the ImageNet dataset. Images found by \megide cannot be considered to match the prompt from a human point of view although a small shape of a dog appears in the center of the image. It should still be noted that these images have higher CLIP scores than natural images that humans would score as perfectly aligned with the score prompt. In a sense, we corroborate previous findings, such as \cite{nguyen2015deep}, proving that neural networks classifiers are easily fooled by evolutionary algorithms or gradient based attacks. Our study highlights the difficulty of directly sampling relevant latent codes in the latent space of a discrete VAE, which is in general bypassed by an additional auto-regressive model that learns to sample meaningful latent codes.

\section{Related Works}

\paragraph{Discrete Optimization} Discrete optimization \cite{parker2014discrete} is a longstanding branch of optimization where some of the variables or constraints are discrete. It is also named combinatorial optimization as the set of feasible solutions is generally finite but evaluation every solution is irrealizable. A first class of algorithms search for an exact solution: in particular, Integer Linear Programming \cite{schrijver1998theory} has been the main subject of interest with exact methods such as cutting plan \cite{gilmore1961linear} or branch and bound \cite{land2010automatic}.  On the other hand, evolutionary algorithms have been very popular as they allow for more flexibility and can be stopped at any iteration to get the best solution explored so far \cite{dorigo1999ant}. Most of those algorithms either incorporate this knowledge by mutating only to feasible solutions or use a relaxation and then project back to the feasible set \cite{lin1992genetic}. For instance, for protein design, exact methods have been proposed for specific biophysical cost functions related to the potential energy \cite{desmet1992dead} of the protein or its binding affinity \cite{ojewole2018bbk}. However, due to the limitations in flexibility of the previous methods, the most popular design tool, incorporated in the popular \textit{Rosetta} library, is based on simulated annealing \cite{huang2011rosettaremodel}. Several work tried to leverage recent breakthroughs of deep learning methods tailored to proteins in the protein design, such as \textit{AlphaDesign} \cite{jendrusch2021alphadesign} or \textit{TrDesign} \cite{norn2021protein}. Other works have proposed gradient-based methods for biological sequence design: some work run gradient ascent on a continuous one-hot representation \cite{liu2020antibody} ; others train an auto-encoder alongside the scoring function and feed the latent representation as input to the latter while gradient is carried out through this latent space \cite{gomez2018automatic}. To the best of our knowledge, our work is the first to propose \qd as a solution for protein design. 

\paragraph{Quality Diversity Optimization}

Searching for novelty instead of quality \cite{lehman2011abandoning} was the first work formulating diversity as an end in itself. It was further refined by introducing a notion of local quality with local competition \cite{lehman2011evolving}. \me \cite{mouret2015illuminating} refined the notion of diversity by introducing the notion of repertoire over a set of descriptors. Further improvement were made on the design the descriptor space such as using a Voronoi tessellation \cite{vassiliades2017using} or use unsupervised descriptors \cite{cully2019autonomous, grillotti2022unsupervised} as defining tailored descriptors can be tedious for some tasks. More efficient ways to illuminate this space have been proposed, such as more efficient mutation operators \cite{vassiliades2018discovering} or covariance matrix adaptation \cite{fontaine2020covariance}. Closest to this work is Differentiable \qd \cite{fontaine2021differentiable} where gradients are used to directly update over the continuous variables. 

Application specific methods were developed to apply \qd algorihtms to noisy domains \cite{flageat2020fast} or when multiple objectives are at stake \cite{pierrot2022multi}. Reinforcement Learning is one of the most popular applications of \qd \cite{arulkumaran2019alphastar} and some methods try to incorporate diversity directly in RL algorithms \cite{nilsson2021policy, pierrot2022diversity}. \qd methods have already been applied to Latent Space Exploration, one of the domains we experiment on. The earliest work was named Innovation Engines \cite{nguyen2015innovation} where authors try to generate diverse images using a \qd approach over ImageNet. Later Latent Space Illumination has been introduced to generate game levels \cite{schrum2020cppn2gan} or images \cite{fontaine2021differentiable}. Our work is also applicable to LSI in the case of a discrete latent space and is the first work of its kind to the best of authors' knowledge.

\section{Conclusion}

We introduced \megide, a genetic algorithm that leverages gradient information over discrete variables for Quality Diversity optimization and show it outperforms classical baselines. Our entropy-based guidance allows to ease the search for good hyperparameters which can be tedious for some applications. Our experiment on protein design is the first using \qd methods to the best of author's knowledge and this opens a way for more applications in the future. Indeed, we believe that generating not only fit but also diverse proteins can be useful as the fitness given by the model does not always translate to good properties in the lab.

Our goal was not to propose a novel image generation technique but while the results of our LSI experiment are interpretable and \megide finds images with higher fitnesses than realistic images, further research can be done to better understand and exploit these results. Our method can be used to create adversarial examples for generative models using discrete latent spaces, that are becoming more and more popular. On the other hand, to generate more realistic images, we could constrain the search to a few latent codes and use a recurrent architecture to generate a more realistic prior, at the cost of potentially facing vanishing or exploding gradients. We leave these open questions for future works.

As our method involves computing gradients of the fitness and descriptor functions, we expect it to be intractable for functions involving large neural networks, such as AlphaFold. Using a surrogate model to guide the search while keeping a large model for evaluation or validation to handle those cases could be a promising research direction.

\section*{Acknowledgement}

This work was supported by the Google's TPU Research Cloud (TRC).

\bibliographystyle{ACM-Reference-Format}
\bibliography{main.bib}

\clearpage

\appendix

\section{Experiment Settings Details}
\label{appendix:exp_details}
In this section we provide additional information about the experiments settings.
\begin{figure*}[b!]
    \centering
    \includegraphics[width=0.99\textwidth]{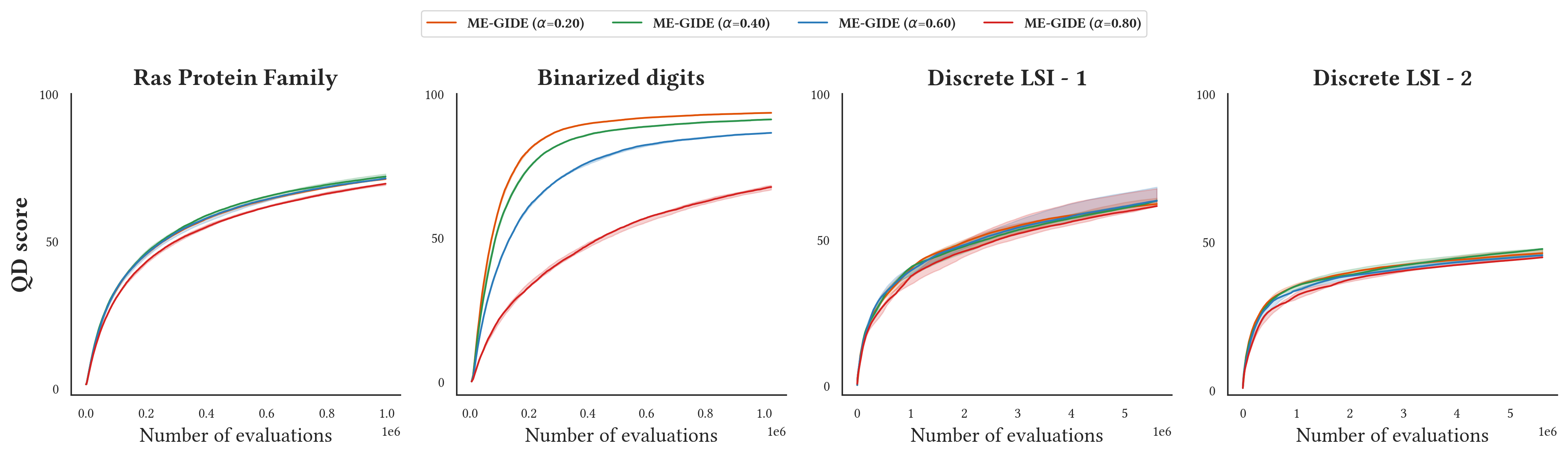}
    \caption{\qd-score evolution on the different domains (median and interquartile range over five seeds) for different values of target entropy. On three out of the four domains, different values of target entropy yield similar results. It demonstrates that the use of a target entropy eases the hyperparameter tuning procedure.}
    \label{fig:target_entropy}
\end{figure*}
\paragraph{Design in the Ras Protein Family}

Input sequences are restricted to a maximum length of $186$ with $25$ possible amino acids. We use the \textit{esm2\_t30\_150M\_UR50D} architecture from the ESM2 repository \footnote{\url{https://github.com/facebookresearch/esm}}, which is made of $30$ attention layers and $150$ millions of parameters in total. We use the $120,000$ elements of the PFAM database to initalize the repertoire in every experiment. We use a batch size of $126$ at each iteration and perform $7937$ iterations to evaluate $1,000,000$ elements in total. Every method is run for a total of five trials. \megide was run for four values of target entropy: $0.2$, $0.4$, $0.6$ and $0.8$. \me was run with different values of number of mutations at each iteration but the best results were obtained with $1-$point mutation at each iteration. \omgmega was run for values of $\sigma_g=0.1,1,10,100,1000$ and the best results were obtained for $\sigma_g=100$. \cmame was run for values of initial standard deviation of $\sigma_0=0.01,0.05,0.1,0.5,1.0,5.0$ and the best results were obtained for $\sigma_0 =0.5$

\paragraph{Binarized digits}

Input images are treated as vectors of length $784$ with $2$ possibilities ($0$ or $1$). We use a RBM with 500 hidden units and we train it with the contrastive divergence algorithm. We initialize the repertoire uniformly at random. We use a batch size of $512$ at each iteration and perform $2000$ iterations to evaluate $1,024,000$ elements in total. Every method is run for a total of five trials. \megide was run for four values of target entropy:  $0.2$, $0.4$, $0.6$ and $0.8$. \me was run with different values of number of mutations at each iteration and with crossover but the best results were obtained with $1-$point mutation at each iteration and no crossover. \omgmegap was run for values of $\sigma_g=1,10,100$  and the best results were obtained for $\sigma_g=10$ in value. \cmame was run for values of initial standard deviation of $\sigma_0=0.01,0.05,0.1,0.5,1.0,5.0$ and the best results were obtained for $\sigma_0 =0.5$

\paragraph{Discrete LSI}

The latent space is made of $32\times32=1024$ codes with $512$ possibilites. We use a VQ-VAE which architecture is detailed in Appendix~\ref{appendix:vq_vae_training}. We initialize the repertoire uniformly at random. We use a batch size of $560$ at each iteration and perform $10000$ iterations to evaluate $5,600,000$ elements in total. Every method is run for a total of five trials. \megide was run for four values of target entropy: $0.2$, $0.4$, $0.6$ and $0.8$. \me was run with different values of number of mutations at each iteration and with crossover but the best results were obtained with $1-$point mutation at each iteration and no crossover. \omgmegap was run for values of $\sigma_g=1,10,100$ and the best results were obtained for $\sigma_g=100$.   \cmame was run for values of initial standard deviation of $\sigma_0=0.01,0.05,0.1,0.5,1.0,5.0$ and the best results were obtained for $\sigma_0 =0.5$. Concerning the descriptors and objective range, the CLIP-based descriptors are scalar values ranging from $0$ to $10$, lower descriptor indicating stronger similarity. To compute the fitness score, we transform the score associated with the fitness prompt by applying the function $x\mapsto(10-x)\times10$. Thus we obtain a score ranging from $0$ to $100$ as displayed in the Figure~\ref{fig:domains}.

\section{Detailed Results for Different Target Entropies}
\label{appendix:full_results}

In all of our experiments we run our method for four values of entropy and only plot the best one in Figure~\ref{fig:main_results}. We show here that for any value of target entropy in $\{0.2,0.4,0.6,0.8\}$, \megide outperforms other baselines on the protein domain and the discrete LSI domains. It highlights the robustness of \megide regarding the exact choice of entropy target. For the binarized digits domain, a lower target entropy induces faster convergence, this is mainly due to the fact that the dimensions of the problem makes it easier to solve, hence favoring methods that can greedily follow gradient signal.

\section{VQ-VAE Architecture}
\label{appendix:vq_vae_training}

To train our VQ-VAE, we follow guidelines of \cite{van2017neural}. We train our VQ-VAE on ImageNet\footnote{https://www.image-net.org/} where images are pre-processed to be of size $128\times128$. We use the same architecture as the authors of the aforementioned paper and use a latent vector of size $32\times32=1024$ with a codebook size of $512$. We use $3$ convolutional layers for the encoder and $3$ layers for the decoder. We use \textit{Adam} optimizer with a learning rate $\eta=2\text{e}-4$ we set the commitment loss coefficient to $\beta=0.25$.

\section{Visualization of the MNIST data in the descriptor space}
\label{appendix:mnist_vizu}

\begin{figure*}[h]
    \centering
    \includegraphics[width=0.5\textwidth]{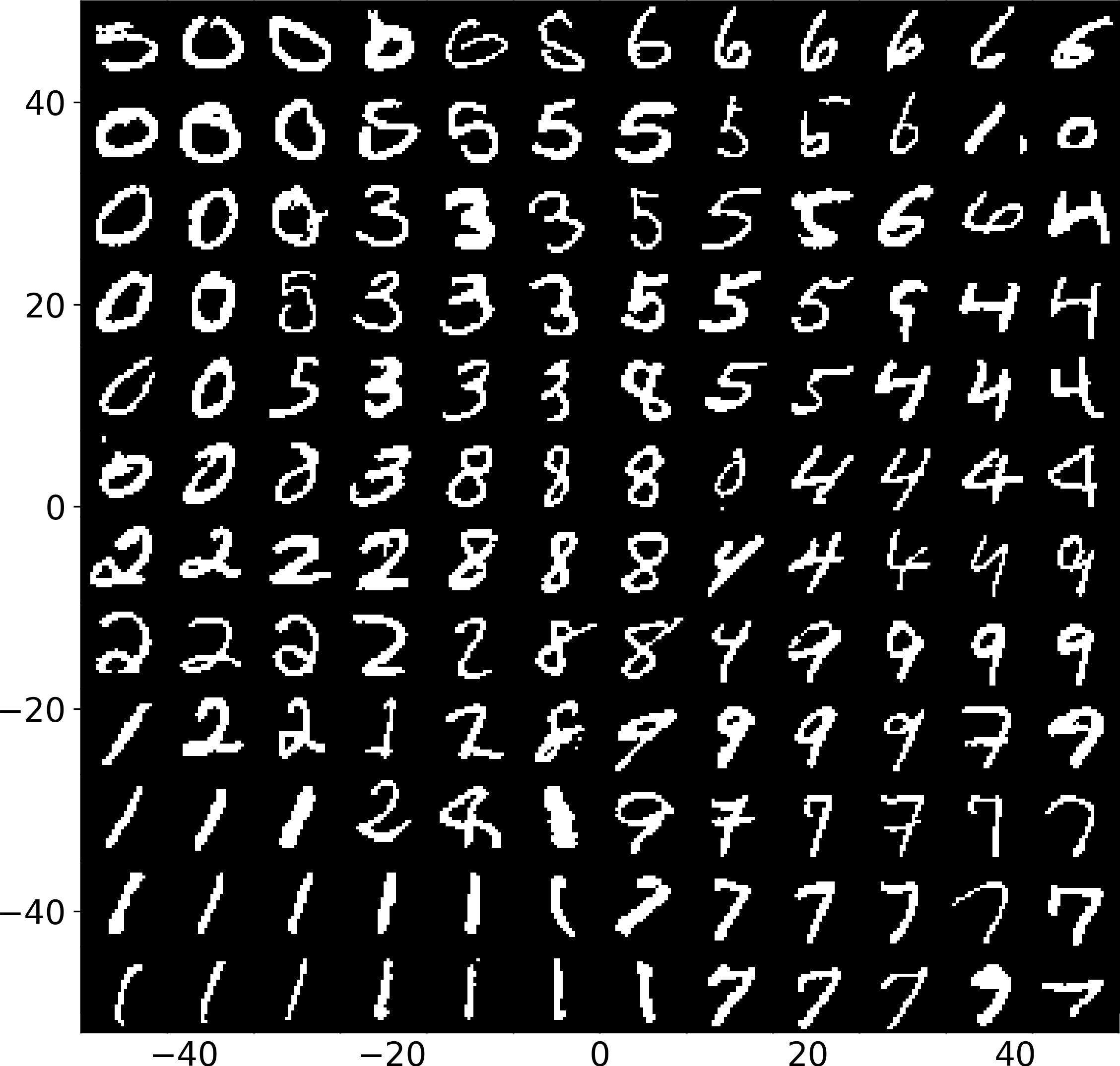}
    \caption{MNIST images projected in a t-SNE reduction of our descriptor space. The whole MNIST dataset is embedded into a the 20-dimensional space defined by the PCA over the hidden units of the RBM. Then we obtain a 2-dimensional representation with t-SNE and we sample images uniformly in this space.}
    \label{fig:mnist_vizualisation}
\end{figure*}

On complex data such as MNIST binary images, it is no easy task to define a relevant descriptor space. To characterize the diversity of different images of digits, one possible solution is to use the features extracted by a Deep Neural Network trained for the classification task. We instead chose to use the features implicitly embedded in the hidden layer of the RBM trained on the MNIST data. As the RBM model is trained for likelihood estimation, we expect it to learn a robust representation of the data, that efficiently separe the different classes of digits. To further validate this choice, we visualize a projection of the MNIST dataset in our embedding space. To do so, we first embed the whole dataset in the 20-dimensional descriptor space, then we project it in dimension 2 using the t-SNE algorithm. In Figure~\ref{fig:mnist_vizualisation} we showcase MNIST images sampled uniformly in the projection space. It demonstrates the fact that our descriptor space properly spread the MNIST different classes and is able to characterize the diversity in the digits' space.

\section{Validation on Protein Data}
\label{appendix:prot_val}

\begin{figure*}[h]
     \centering
     \begin{subfigure}[b]{0.4\textwidth}
         \centering
         \includegraphics[width=\textwidth]{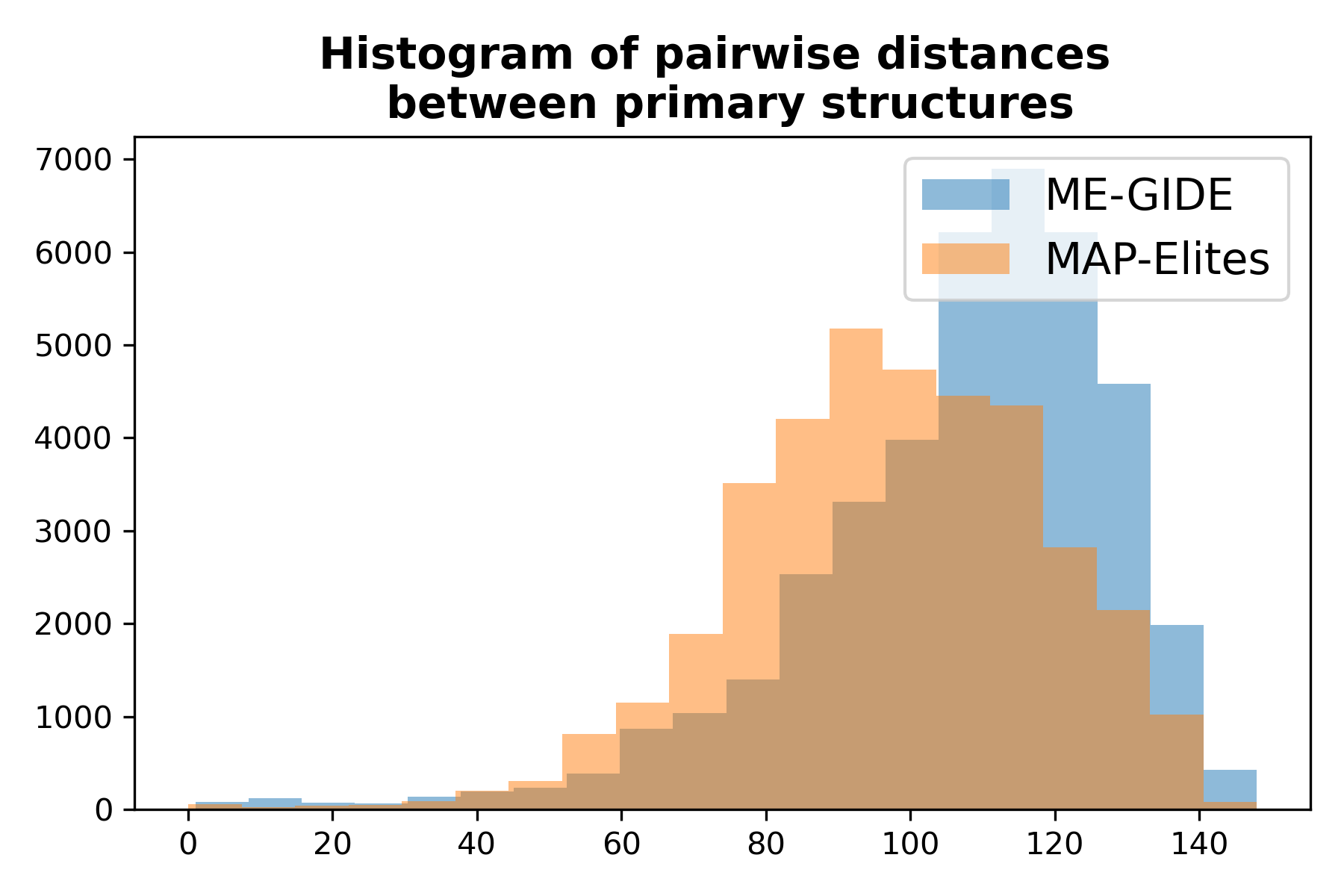}
         \caption{Histogram of edit distances between sequence data over the \megide and \me repertoires.}
     \end{subfigure}
     \hfill
     \begin{subfigure}[b]{0.4\textwidth}
         \centering
         \includegraphics[width=\textwidth]{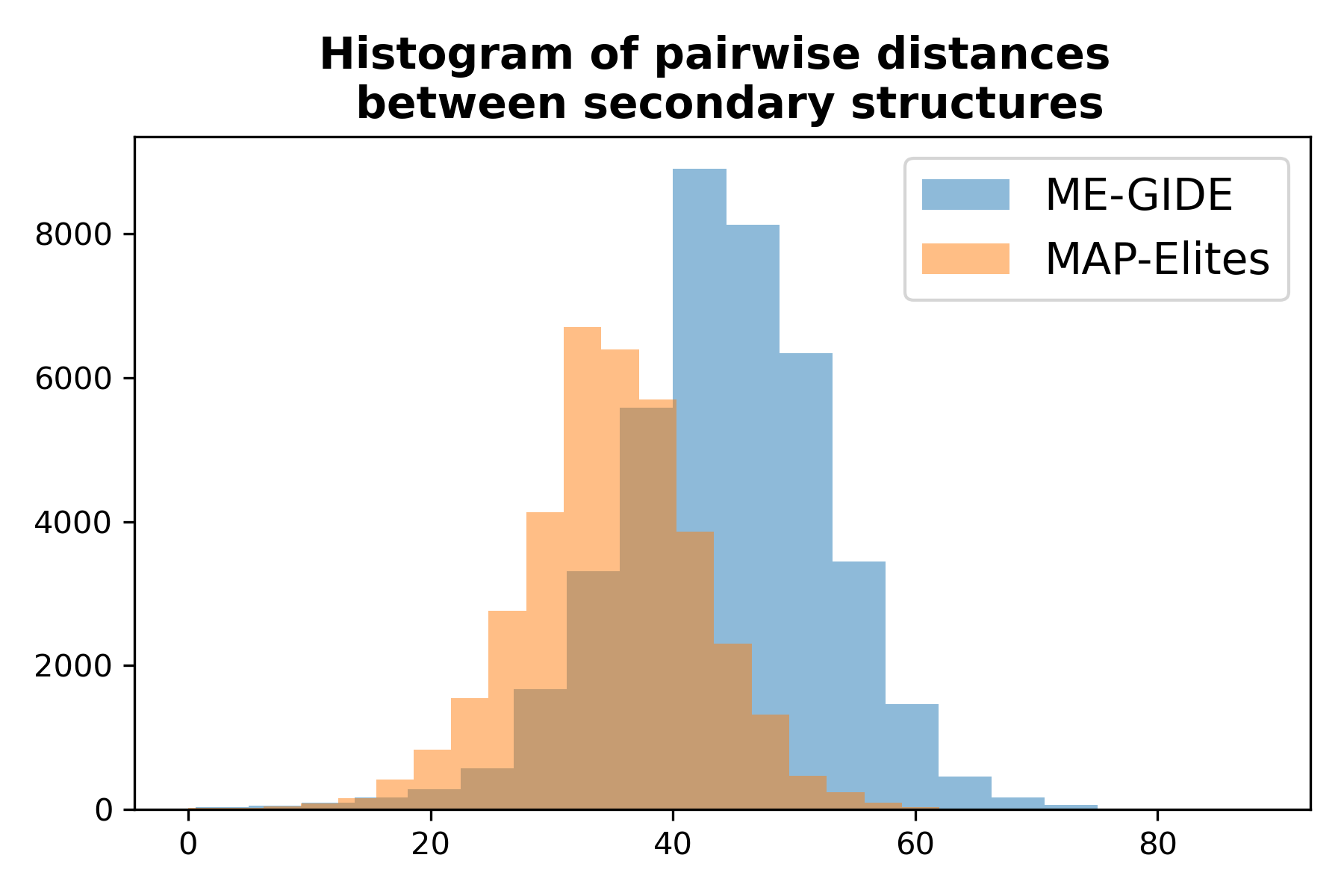}
         \caption{Histogram of edit distances between secondary structure over the \megide and \me repertoires.}
     \end{subfigure}
    \caption{\megide finds more diverse solutions in the sequence space.}
    \label{fig:validation_proteins}
\end{figure*}

We visualize the diversity obtained on proteins with two information: primary structure and secondary structure. First, we sub-sample to $300$ the repertoire obtained with \me and \megide by performing a $K=300$-means clustering on the centroids of the repertoire. Then every protein from the original repertoire is added to the new one, keeping only the most fit in every region. We evaluate diversity in the primary structure (amino acid sequence) by computing the pairwise Levenshtein distances between each protein. We display the histograms of the distributions of pairwise distances bewtween \me and \megide in Figure~\ref{fig:validation_proteins}. 

\clearpage

\section{Pseudo-code for the baselines}
\label{app:baselines}

\input{me.tex}
\input{omgmegap.tex}

\input{cmamep.tex}

\end{document}

%% file: math_commands.tex
\usepackage{amsmath,amsfonts,bm}

\def\eqref#1{equation~\ref{#1}}
\def\1{\bm{1}}

\DeclareMathAlphabet{\mathsfit}{\encodingdefault}{\sfdefault}{m}{sl}
\SetMathAlphabet{\mathsfit}{bold}{\encodingdefault}{\sfdefault}{bx}{n}

\DeclareMathOperator*{\argmax}{arg\,max}

%% file: mycommands.tex
\usepackage{xspace}
\usepackage{units}

\definecolor{myred}{rgb}{0.8,0,0}
\definecolor{mygreen}{rgb}{0,0.6,0}
\definecolor{myblue}{rgb}{0,0,0.7}

\newcommand{\megide}{{\sc me-gide}\xspace}
\newcommand{\me}{{\sc map-elites}\xspace}

\newcommand{\cmame}{{\sc cma-map-elites}\xspace}

\newcommand{\qd}{{\sc qd}\xspace}
\newcommand{\gide}{{\sc gide}\xspace}

\newcommand{\omgmega}{{\sc omg-mega}\xspace}
\newcommand{\omgmegap}{{\sc omg-mega (proj.)}\xspace}
\newcommand{\cmamep}{{\sc cma-map-elites (proj.)}\xspace}
\newcommand{\cmamega}{{\sc cma-mega}\xspace}

%% file: gide.tex
\makeatletter
\newcommand{\removelatexerror}{\let\@latex@error\@gobble}
\makeatother

\begin{figure}[h!]
\removelatexerror
\centering
\begin{algorithm}[H]
    \small
    \SetAlgoLined
    \DontPrintSemicolon
    \SetKwInput{KwInput}{Given}
    \KwInput{a batch of $B$ elements $x_n$ from the repertoire and their associated gradients $\nabla_n$.
    }
    \For{$1 \leq n \leq B$}{
    \texttt{\\}
    \tcp{Compute the distribution}
    Compute $\mathbf{\tilde{\delta}_n}$ using Equation~\ref{eq:appx_diff}.\;
    Compute the proposal distribution $p_n$ using Equation~\ref{eq:probs}.\;
    \texttt{\\}
    \tcp{Target entropy}
    Adjust $T$ to obtain a target entropy $\bar H =\bar H_\text{target}$ over the proposal distributions.\;
    \texttt{\\}
    \tcp{Sample}
    Draw a mutant $x'_n$ from each $x_n$ according to $p_n$.\;
    \caption{Gradient Informed Discrete Emitter (\gide)}
    \label{alg:gide}
    }
    Return ${x'_1, ..., x'_B}$

\end{algorithm}

\end{figure}

%% file: me_gide.tex
\makeatletter
\makeatother

\begin{figure}[h!]
\removelatexerror
\centering
\begin{algorithm}[H]
    \small
    \SetAlgoLined
    \DontPrintSemicolon
    \SetKwInput{KwInput}{Given}
    \KwInput{ the number of cells $M$ and tesselation $\mathcal{T}$ ; the batch size $B$ and the number of iterations $N$ ;the descriptors function $\mathbf{c}$ and the multi-objective function $\mathbf{f}$ ; a target entropy value $H_\text{target}$ ; the initial population of solution candidates $\{\mathbf{x}_k\}$}
    \texttt{\\}
    
    \tcp{Initialization}
    For each initial solution, find the cell corresponding to its descriptor and add initial solutions to their cells.\;
    \texttt{\\}
    
    \tcp{Main loop}
    \For{$1 \leq n_{\text{steps}} \leq N$}{
    \texttt{\\}
        \tcp{Select new generation}
        Sample uniformly $B$ solutions $x_n$ in the repertoire with replacement\;
        \texttt{\\}\tcp{Compute the gradients}
        Randomly draw weights $\mathbf{w^{(n)}} \sim \mathcal{N}(0, I)$\;
        Compute gradients $\nabla_{x_n}f(x_n)$ and $\nabla_{x_n}c(x_n)$\;
        Normalize the gradients.\;
        Compute combined updates $\nabla_n = |w^{(n)}_0| \nabla_{x_n}f(x_n) + \sum\limits_{i=1}^n w^{(n)}_i \nabla_{x_n}c_i(x_n)$ \;
        
        \tcp{Gradient-Informed Discrete Emitter}
        Use Algorithm~\ref{alg:gide} to sample mutants $x'_n$\;
        \texttt{\\}
        \tcp{Addition in the archive}
        Add each mutant in its corresponding cell if it improves the cell fitness, otherwise discard it.\;
        \texttt{\\}
    }
    
    \caption{MAP-Elites with Gradient Informed Discrete Emitter (\megide)}
    \label{alg:pseudocode}
\end{algorithm}

\end{figure}

%% file: me.tex
\makeatletter
\makeatother

\removelatexerror
\centering
\begin{algorithm}[H]
    \small
    \SetAlgoLined
    \DontPrintSemicolon
    \SetKwInput{KwInput}{Given}
    \KwInput{the number of cells $M$ and tesselation $\mathcal{T}$; the batch size $B$ and the number of iterations $N$; the descriptors function $\mathbf{c}$ and the multi-objective function $\mathbf{f}$.}
    \texttt{\\}
    
    \tcp{Initialization}
    For each initial solution, find the cell corresponding to its descriptor and add initial solutions to their cells.\;
    \texttt{\\}
    
    \tcp{Main loop}
    \For{$1 \leq n_{\text{steps}} \leq N$}{
    \texttt{\\}
    
        \tcp{Select new generation}
        Sample uniformly $B$ solutions $x_n$ in the repertoire with replacement\;
        \texttt{\\}

        \tcp{Randomly mutate}
        Sample mutants $x'_n$ by randomly flipping one dimension.
        \texttt{\\}
        
        \tcp{Addition in the archive}
        Add each mutant in its corresponding cell if it improves the cell fitness, otherwise discard it.\;
        \texttt{\\}
    }
    
    \caption{\me}
    \label{alg:map-elites}
\end{algorithm}

%% file: omgmegap.tex
\makeatletter
\makeatother

\removelatexerror
\centering
\begin{algorithm}[H]
    \small
    \SetAlgoLined
    \DontPrintSemicolon
    \SetKwInput{KwInput}{Given}
    \KwInput{the number of cells $M$ and tesselation $\mathcal{T}$; the batch size $B$ and the number of iterations $N$ ;the descriptors function $\mathbf{c}$ and the multi-objective function $\mathbf{f}$; the initial population of solution candidates $\{\mathbf{x}_k\}$; the standard deviation of the gradient weights $\sigma_g$;}
    \texttt{\\}
    
    \tcp{Initialization}
    For each initial solution, find the cell corresponding to its descriptor and add initial solutions to their cells.\;
    \texttt{\\}
    
    \tcp{Main loop}
    \For{$1 \leq n_{\text{steps}} \leq N$}{
    \texttt{\\}
    
        \tcp{Select new generation}
        Sample uniformly $B$ solutions $x_n$ in the repertoire with replacement\;
        \texttt{\\}
        
        \tcp{Compute the gradients}
        Randomly draw weights $\mathbf{w^{(n)}} \sim \mathcal{N}(0, \sigma_g I)$\;
        Compute gradients $\nabla_{x_n}f(x_n)$ and $\nabla_{x_n}c(x_n)$\;
        Normalize the gradients.\;
        Compute combined updates $\nabla_n = |w^{(n)}_0| \nabla_{x_n}f(x_n) + \sum\limits_{i=1}^n w^{(n)}_i \nabla_{x_n}c_i(x_n)$ \;
        
        \tcp{OMG-MEGA emitter}
        $x'_n = x_n + \nabla_n$
        \texttt{\\}
        
        \tcp{Project on the discrete space.}
        $x'_n = \text{proj}(x'_n)$

        \tcp{Addition in the archive}
        Add each mutant in its corresponding cell if it improves the cell fitness, otherwise discard it.\;
        \texttt{\\}
    }
    
    \caption{Projected Objective and Measure Gradient MAP-Elites via Gradient Arborescence (\omgmegap)}
    \label{alg:pseudocode}
\end{algorithm}

%% file: cmamep.tex
\makeatletter
\makeatother

\removelatexerror
\centering
\begin{algorithm}[H]
    \small
    \SetAlgoLined
    \DontPrintSemicolon
    \SetKwInput{KwInput}{Given}
    \KwInput{the number of cells $M$ and tesselation $\mathcal{T}$; the batch size $B$ and the number of iterations $N$; the descriptors function $\mathbf{c}$ and the multi-objective function $\mathbf{f}$; the initial population of solution candidates $\{\mathbf{x}_k\}$; the population of emitters $E$; the initial standard deviation of the covariance matrices $\sigma_0$.}
    
    \texttt{\\}
    
    \tcp{Initialization}
    Initialize CMA emitters with the parameter $\sigma_0$.
    For each initial solution, find the cell corresponding to its descriptor and add initial solutions to their cells.\;
    \texttt{\\}
    
    \tcp{Main loop}
    \For{$1 \leq n_{\text{steps}} \leq N$}{
    \texttt{\\}
    
        \tcp{Select new generation by sampling $B$ solutions $(x'_n)_{1\leq i\leq B}$ using CMA emitters}
        \For{$1 \leq i \leq B$}{
            Select emitter $e_i$ from $E$ which has generated the least solutions out of all emitters in E.
            
            Unpack mean $m_i$ and covariance matrix $C_i$ from $e_i$.
            
            Draw a candidate from the emitter $(x'_n)_i 
            \sim \mathcal{N}(m_i, C_i)$.
        }
        \texttt{\\}
        \tcp{Project on the discrete space.}
        $x'_n = \text{proj}(x'_n)$

        \tcp{Addition in the archive}
        Add each mutant in its corresponding cell if it improves the cell fitness, otherwise discard it.\;
        \texttt{\\}

        \tcp{Update of the emitters}
        Update the parameters of the CMA emitters, their mean $m$ and their covariance matrix $C$.
    }
    
    \caption{Projected Covariance Matrix Adaptation MAP-Elites (\cmamep)}
    \label{alg:pseudocode}
\end{algorithm}